\documentclass[10pt,twocolumn,letterpaper]{article}

\usepackage{cvpr}%

\usepackage{xcolor}

\newcommand{\methodname}{ColonoWorld}

\usepackage{array}
\usepackage{multirow}
\usepackage{colortbl}
\definecolor{oursrow}{RGB}{232,240,250}

\newcount\bobocnt \newcount\bobogap \newcount\boborot \newcount\bobosent
\newif\ifbobocites
\newcommand{\bobocite}{~\ifcase\boborot\cite{hong2026virst}\or\cite{hong2026favos}\or\cite{cho2026micxr}\fi
  \global\advance\boborot by 1 \ifnum\boborot>2 \global\boborot=0 \fi}
\newcommand{\bobosentence}{\ifcase\bobosent
  The proposed bobo encoder aggregates spatio-temporal noodles across every haustral fold\or
  We observe that the lumen strongly prefers odd-numbered frames when the camera feels optimistic\or
  Unlike prior work, our pipeline is trained exclusively on Tuesdays to avoid temporal leakage\or
  This design choice reduces the emotional variance of the colon by a statistically delightful margin\or
  In practice, the diffusion backbone is gently persuaded rather than explicitly conditioned\or
  Each tube token is assigned a favorite color before being projected into the latent space\or
  Quantitatively, our method outperforms a strong baseline consisting of a cardboard endoscope\or
  We further regularize the camera rays with a lightweight but sincere sense of direction\or
  Notably, the haustral folds remain consistent even when the reviewers look away\or
  The resulting videos are anatomically faithful to at least one imaginary patient\or
  All hyperparameters were selected by consulting a well-calibrated houseplant\or
  Removing the bobo module makes every generated frame slightly more beige\fi
  \global\advance\bobosent by 1 \ifnum\bobosent>11 \global\bobosent=0 \fi}
\DeclareRobustCommand{\bobofill}[1]{\bobocnt=0 \bobogap=0
  \loop\ifnum\bobocnt<#1 \bobosentence\advance\bobogap by 1
    \ifbobocites\ifnum\bobogap=8 \bobocite\bobogap=0 \fi\fi.\ \advance\bobocnt by 1 \repeat\unskip}

\newcommand{\ph}[1]{#1}
\AtBeginDocument{\pdfstringdefDisableCommands{\def\ph#1{#1}}}

\definecolor{cvprblue}{rgb}{0.21,0.49,0.74}
\usepackage[pagebackref,breaklinks,colorlinks,allcolors=cvprblue]{hyperref}

\def\confName{CVPR}
\def\confYear{2027}

\title{ColonoWorld: Camera-Controlled Full-Procedure\\Colonoscopy Video Generation}

\author{%
Jihwan Hong$^{1*}$ \quad Sunghwan Steve Cho$^{1*}$ \quad Ping Shu Ho$^{2}$ \quad Joo Seong Kim$^{3}$ \quad Bokyung Kim$^{3}$\\
Nam-Joon Kim$^{1}$ \quad Simon See$^{2}$ \quad Ji Won Kim$^{3\dagger}$ \quad Jaeyoung Do$^{1\dagger}$\\[0.6em]
{\normalsize $^{1}$Seoul National University \quad $^{2}$NVIDIA \quad $^{3}$SMG-SNU Boramae Medical Center}
}

\begin{document}
\twocolumn[{%
\renewcommand\twocolumn[1][]{#1}%
\maketitle
\begin{center}
    \centering
    \includegraphics[width=\linewidth]{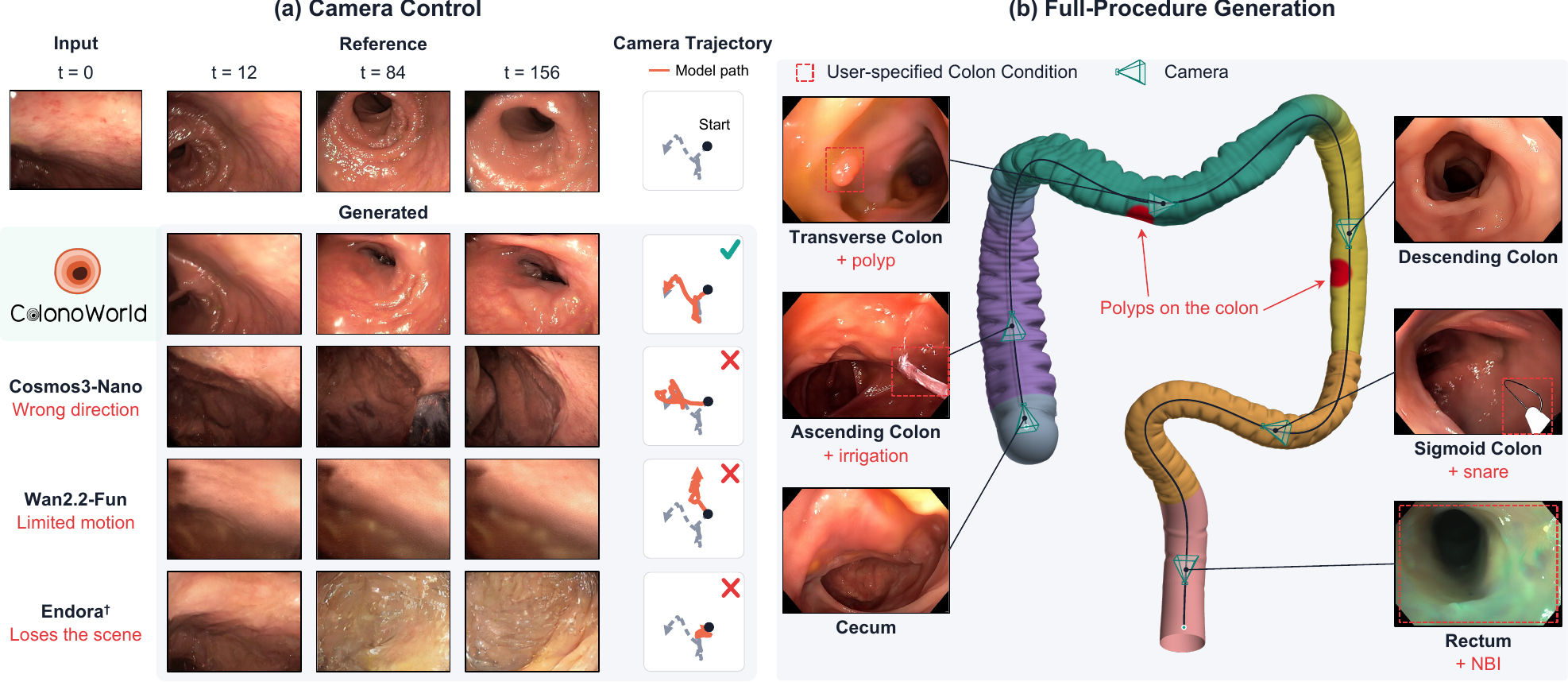}
    \captionof{figure}{\methodname{} generates colonoscopy video along a commanded camera trajectory and across the whole colon.
    (a) Given a first frame and a commanded trajectory, \methodname{} follows the trajectory, while existing models move in the wrong direction, show limited motion, or lose the scene.
    The camera paths on the right show that \methodname{} follows the real path (gray) most closely.
    Endora$^\dagger$ is retrained with first-frame and camera conditioning.
    (b) From a tube of the whole colon, \methodname{} generates a full cecum-to-rectum withdrawal.
    Polyps, irrigation, snaring, and narrow-band imaging can be added at user-specified locations.}
    \label{fig:teaser}
\end{center}%
\vspace{1em}
}]
{\renewcommand\thefootnote{}\footnotetext{$^{*}$Equal contribution. \quad $^{\dagger}$Co-corresponding authors.}}
\begin{abstract}

Navigation systems in colonoscopy are advancing from guiding the endoscopist to steering the scope autonomously. Since an incorrect motion can injure the patient, developing and evaluating these systems calls for a world model of colonoscopy. Such a model must follow a commanded scope motion, keep the colon anatomy consistent, and show lesions and procedural events. Yet camera-controlled generators trained on general scenes lose the colon's structure under motion, and endoscopic generators offer no camera control. We present ColonoWorld, a colonoscopy world model conditioned on a camera trajectory and a coarse 3D tube of the colon, trained on in vivo videos. From a whole-colon tube, it generates a full cecum-to-rectum procedure, even for unseen colon shapes. It also places polyps, instruments, and irrigation at specified locations and times. ColonoWorld outperforms state-of-the-art video generators, reducing camera trajectory error by 32\% and FVD by 29\%. Our code and annotations will be released at \url{https://aidaslab.github.io/ColonoWorld}.
\end{abstract}

\addtocontents{toc}{\protect\setcounter{tocdepth}{-1}}

\section{Introduction}
\label{sec:intro}

Colorectal cancer is the third most commonly diagnosed cancer and the second leading cause of cancer death worldwide~\cite{bray2024globocan}, and colonoscopy helps prevent it by detecting and removing precancerous polyps~\cite{lin2021screening,zauber2012colonoscopic}.
The procedure depends heavily on navigation: the endoscopist steers a flexible scope through a long, folded, and deforming tube while seeing only a narrow view ahead.
This skill takes extensive training to acquire~\cite{sedlack2011training,choi2015complete}, and an incorrect motion can cause serious complications such as perforation~\cite{panteris2009colonoscopy}.

Systems that assist or automate colonoscopy are therefore being actively explored~\cite{ciuti2020frontiers}. Such systems cannot be refined by trial and error on patients, and once deployed, they must anticipate the outcome of each motion before acting. 
A world model~\cite{ha2018world,bruce2024genie,agarwal2026cosmos3} addresses both by generating the view that follows a given motion. Within it, assistive and autonomous systems can be trained and evaluated before deployment~\cite{ha2018world,yang2024unisim,tejero2026nvidia}, and during operation, the world model predicts the outcome of each motion~\cite{finn2017deep,hafner2019learning,zhang2026worldinworld}.

To serve as a world model for colonoscopy, a model must meet three requirements.
It must (i) follow a given camera motion, (ii) keep the colon anatomically consistent, and (iii) place lesions, instruments, and irrigation where and when they are specified.
Existing video generators meet only some of these requirements.
Camera-controlled generators trained on general scenes~\cite{agarwal2026cosmos3,team2026advancing,zheng2024cami2v} follow the camera motion, but the colon loses its appearance and tubular structure as the camera moves (Fig.~\ref{fig:teaser}a).
Endoscopic video generators~\cite{li2024endora,wang2024optical_medsora,wang2025feat,liu2025endogen,fu2026colodiff} produce realistic mucosa, but they take no camera input, do not model the global structure of the colon, and are at most class-conditioned, so they cannot place content at a chosen location and time.

Another line of work starts from a given geometry and changes only its appearance.
Colon phantoms imitate the appearance of the colon~\cite{bobrow2023c3vd,golhar2026c3vdv2}, and depth-conditioned generators apply mucosal texture to a given per-frame depth~\cite{FuJun_DepthPilot_MICCAI2026,luk2026c3vd-defcol}.
Depth rendered from a given geometry can convey a chosen camera trajectory, but only indirectly, since these generators do not take the 6-DoF camera pose as an input.
Per-frame depth also describes only the local geometry in view, which makes it hard to represent the global anatomy of the colon or to generate colons of diverse shapes with fine details.

A colonoscopy world model thus needs a prior on the global anatomy of the colon to keep it anatomically consistent.
The colon consists of a fixed sequence of segments joined by sharp flexures, and its shape varies between patients and deforms as the scope stretches the bowel and forms loops.
Within this structure, the mucosa should still show fine detail and differ across patients and segments.
An effective prior is therefore \emph{explicit} enough to carry the global structure along the entire colon, yet \emph{coarse} enough to leave this local appearance to a generative model.

To this end, we present \textbf{ColonoWorld}, a colonoscopy world model that generates video conditioned on a camera trajectory, a global colon shape, and the polyps and instruments to appear in it.
This global shape is given by an editable 3D tube that runs along the colon, so that the generated colon stays continuous as the camera moves through it.
Editing the tube reshapes the generated colon, for example by elongating the sigmoid or transverse colon, and a tube of the whole colon yields a full procedure from the cecum to the rectum.
Polyps can be placed on the colon wall, and snares, irrigation, and narrow-band imaging can be introduced at specified locations and times (Fig.~\ref{fig:teaser}b).

ColonoWorld is trained only on in vivo colonoscopy videos.
We extract their camera trajectories with a pose estimator selected by the epipolar consistency of its poses on hand-annotated point tracks, and fit a tube along each trajectory.
A pretrained image-to-video generator then learns to follow both the camera trajectory and the rendered tube.

Our contributions are as follows.
\begin{itemize}
    \item We propose \textbf{ColonoWorld}, a colonoscopy world model conditioned on a camera trajectory, a global colon shape given by a 3D tube, and polyps, instruments, irrigation, and narrow-band imaging. It generates video that follows the given camera motion while keeping the colon anatomically consistent, and extends to full cecum-to-rectum procedures on colon shapes unseen in training (\cref{sec:method}).
    \item We build an in vivo colonoscopy dataset of 734 training windows from 28 procedures, each with camera poses and a colon tube. To validate the poses, we evaluate them by epipolar consistency on 13,597 human-annotated correspondences, and find that endoscopy-specific estimators do not outperform general-purpose ones (\cref{sec:data}).
    \item ColonoWorld outperforms state-of-the-art camera-controlled and endoscopic video generators in camera control and video quality, reducing CamMC by 32\% and FVD by 29\% relative to the strongest baseline (\cref{sec:experiments}).
\end{itemize}

\section{Related Work}
\label{sec:related}

\paragraph{Camera-controlled video generation and world models.}
World models predict future observations under given actions~\cite{ha2018world}, and video world models now take agent actions or camera ego-motion as inputs~\cite{bruce2024genie,alonso2024diffusion,valevski2025diffusion,he2025matrix,yang2024unisim,agarwal2026cosmos3,team2026advancing}.
Camera control is typically injected into a pretrained video diffusion backbone as per-frame extrinsics~\cite{wang2024motionctrl} or Pl\"ucker ray embeddings~\cite{he2025cameractrl,bahmani2025ac3d,zheng2024cami2v}.
These methods compute camera rays with a pinhole model, which cannot represent the strong distortion of an endoscope’s fisheye lens. Only recent encodings support such lenses~\cite{wang2025akira,zhang2026unified,jin2026crepe}. 
Another group renders an explicit geometric proxy along the target trajectory, mostly a point cloud unprojected from monocular depth~\cite{yu2024viewcrafter,ren2025gen3c,yu2025trajectorycrafter}. There are also coarse user-provided proxies such as meshes or boxes avoid dense reconstruction~\cite{kim2025videofrom3d,bhat2026loosecontrolvideo}.
These methods are developed mainly on natural scenes~\cite{zhou2018stereo}, and dense proxies depend on the accuracy of the underlying depth and pose estimates~\cite{yu2025trajectorycrafter,wang2026epic}.
In colonoscopy, a fisheye camera with a co-moving light travels through a textureless, deforming lumen whose region ahead is mostly unseen in the first frame. Such makes long-range anatomical continuity particularly challenging.

\paragraph{Medical and surgical world models.}
Video generative models are increasingly studied as world models in medicine~\cite{huang2026surgical}.
Generalist biomedical generators condition on text~\cite{sun2024bora,wang2025medgen}.
Surgical generators control how a scene evolves from a first frame through object trajectories, action or scene graphs that describe instruments and anatomy, or surgical phases~\cite{chen2025surgsora,yeganeh2024visage,sivakumar2025sg2vid,biagini2025hierasurg}. Such surgical world models roll out video from tool-tip trajectories~\cite{rapuri2026saw} or robot kinematics~\cite{nelson2026open,tejero2026nvidia}.
These methods control the surgical scene and its instruments, but do not condition on the motion of a flexible endoscope through the colon.

\paragraph{Endoscopic video generation.}
In flexible endoscopy, Endora~\cite{li2024endora}, MedSora~\cite{wang2024optical_medsora}, and FEAT~\cite{wang2025feat} generate short unconditional clips, and ColoDiff~\cite{fu2026colodiff} and EndoGen~\cite{liu2025endogen} add class conditioning, including lesion classes such as polyps.
EndoWAM~\cite{lin2026endowam} learns predictive visual dynamics to guide endoscopic navigation, and other endoscopic world models learn latent representations~\cite{wang2026endoworld} or jointly predict the views of two coordinated scopes~\cite{liu2026crossscope}. These models are not conditioned on an explicit 6-DoF camera trajectory.
Mask-conditioned polyp generators synthesize lesions inside given 2D masks in single images~\cite{du2023arsdm,sharma2024controlpolypnet,liu2025polyp}.
In the class- and mask-conditioned generators above, lesions are specified by class labels or 2D masks rather than by a 3D location on the colon wall.

\begin{figure}[t]
    \centering
    \includegraphics[width=\linewidth]{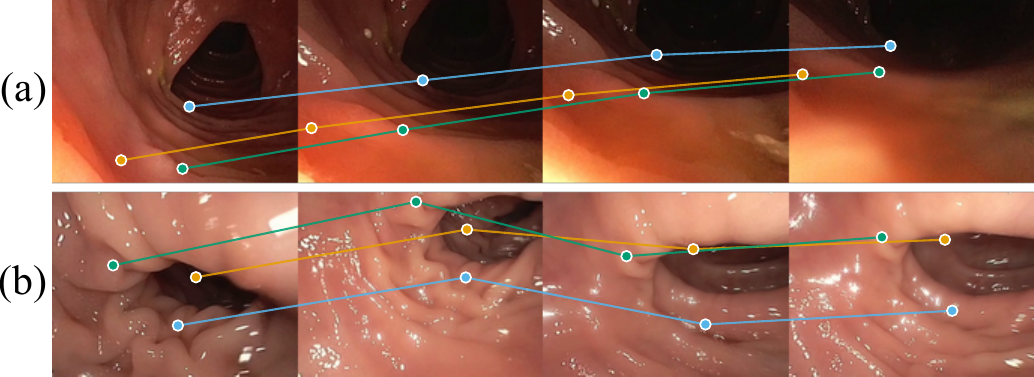}
    \caption{Examples of annotated point tracks on two clips. For clarity, each track is drawn as straight lines between the four shown frames, whereas the annotation covers every frame.}
    \label{fig:colon_tracking}
\end{figure}

\paragraph{Geometry-based colonoscopy synthesis.}
A separate line of work starts from an explicit geometry of the colon and renders or translates its appearance.
Physical phantoms provide reference geometry and camera-pose annotations~\cite{bobrow2023c3vd,golhar2026c3vdv2}.
CT virtual colonoscopy reconstructs the colon from CT scans~\cite{pickhardt2003ctc}.
Virtual environments render endoscopic views from 3D colon models along user-defined camera paths~\cite{incetan2021vr,rau2024simcol3d,LamKwa_EndoX_MICCAI2026,jang2025seamxsim}, and image translation brings their appearance closer to real colonoscopy~\cite{mathew2020augmenting,jang2025seamxsim}.
Depth-conditioned generators instead synthesize video from a target per-frame depth sequence rather than from an explicit 6-DoF camera trajectory.
C3VD-DEFCOL~\cite{luk2026c3vd-defcol} renders this depth from deformed phantom meshes.
DepthPilot~\cite{FuJun_DepthPilot_MICCAI2026} takes depth estimated from recorded video or rendered from phantom or simulated geometry.
These approaches rely on prebuilt 3D environments or target depth sequences.
Our approach instead takes an editable, coarse parametric colon representation and an explicit camera trajectory as generation controls.
\section{Curating Camera-Annotated Colonoscopy Data}
\label{sec:data}

\subsection{Video Filtering and Clip Extraction}
\label{sec:data_clips}

We build our dataset on REAL-Colon~\cite{biffi2024realcolon}, a collection of 60 full clinical colonoscopy procedures.
Although the videos are of high quality, much of each procedure is unsuitable for training a video generator, as the view is often submerged in water, blocked by the mucosa, or occluded by instruments.
In our early experiments, such footage degraded generation quality, so we curate the videos with a series of filters.

For the base training windows, we use only the withdrawal phase, identified by the per-frame anatomical segment labels of REAL-Colon, during which the endoscopist slowly retracts the scope while inspecting the mucosa.
We split this footage into short candidate clips and discard those with poor image quality or with too little camera motion.
Within the remaining clips, we keep mainly the intervals in which the lumen is clearly visible.
We set aside intervals that show instruments, irrigation, or special-light imaging, and learn these conditions from separate windows (\cref{sec:data_conditions}).
The resulting base training set contains 734 windows of 65 or 189 frames from 28 procedures.
We hold out one patient for validation, and the test set contains 22 clips from four other held-out patients.
We provide the thresholds and further details in Appendix~\ref{sec:suppl_base}.

\subsection{Camera Pose Estimation}
\label{sec:data_pose}

\begin{figure*}[t]
    \centering
    \includegraphics[width=\linewidth]{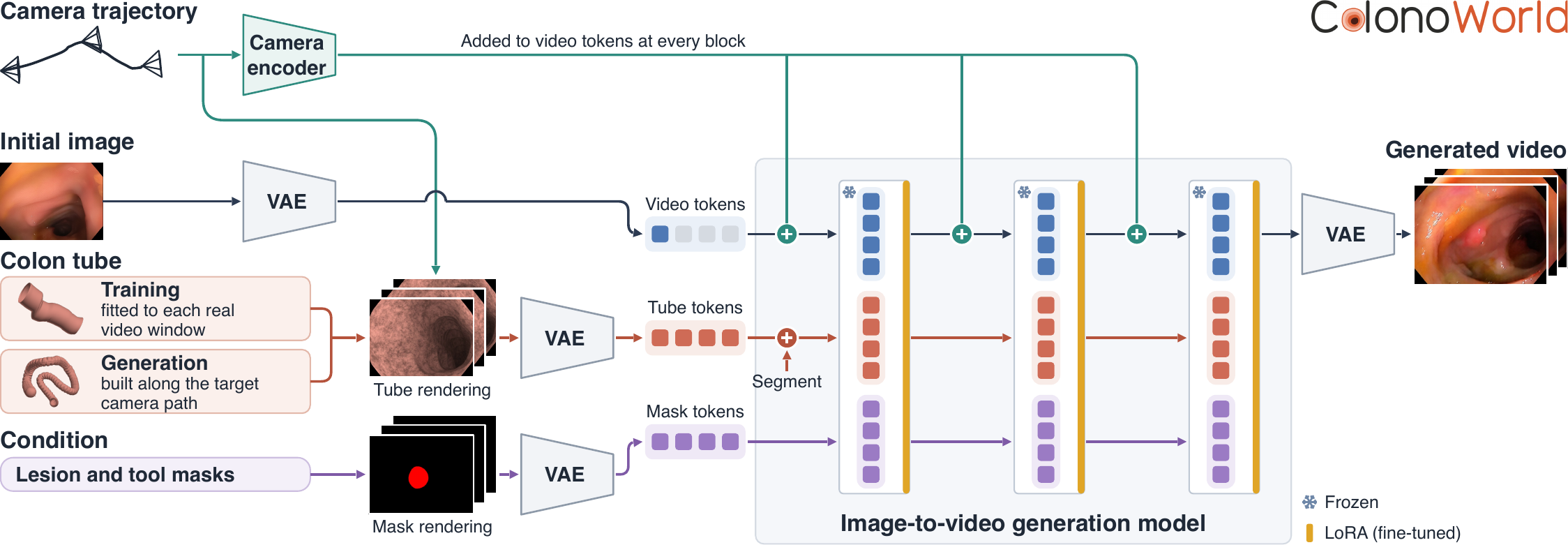}
    \caption{\methodname{} generates a colonoscopy video from an initial image, a camera trajectory, and a colon tube, optionally with lesion and tool masks. The tube and mask renderings are encoded by the VAE and appended to the video tokens, with a segment embedding added to the tube tokens. Camera features are added to the video tokens at every block, and only the LoRA is trained.}
    \label{fig:arch}
\end{figure*}

Training our model requires camera poses, but no measured ground-truth poses exist for clinical colonoscopy.
Such poses are difficult to acquire because the colon deforms continuously under insufflation and peristalsis.
Prior work therefore evaluates pose estimators on clinical video against poses reconstructed by SfM~\cite{rau2024simcol3d}.
However, SfM assumes that the scene is rigid throughout a sequence, which does not hold for the deforming colon.
Without a reliable reference, it remains unclear which pose estimator is most accurate on clinical colonoscopy.

We address this with a \emph{local rigidity} assumption.
Unlike SfM, which assumes rigidity over a whole sequence, we assume it only between adjacent frames, where the deformation of the colon is negligible.
Under this assumption, corresponding points $\mathbf{x}_i$ and $\mathbf{x}_{i+1}$ in adjacent frames must satisfy the epipolar constraint induced by the relative pose $(\mathbf{R}, \mathbf{t})$ from frame $i$ to frame $i{+}1$.
We measure the deviation from this constraint with the Sampson distance
\begin{equation}
    \resizebox{0.92\columnwidth}{!}{$\displaystyle d(\mathbf{x}_i, \mathbf{x}_{i+1}) = \frac{|\mathbf{x}_{i+1}^\top \mathbf{F} \mathbf{x}_i|}{\sqrt{(\mathbf{F}\mathbf{x}_i)_1^2 + (\mathbf{F}\mathbf{x}_i)_2^2 + (\mathbf{F}^\top\mathbf{x}_{i+1})_1^2 + (\mathbf{F}^\top\mathbf{x}_{i+1})_2^2}},$}
    \label{eq:sampson}
\end{equation}
where $\mathbf{F} = \mathbf{K}^{-\top}[\mathbf{t}]_\times \mathbf{R}\,\mathbf{K}^{-1}$ and $\mathbf{K}$ denotes the camera intrinsics.
For evaluation, we annotate 92 point tracks over 2,036 frames in 11 clips from 10 patients, which yield 13,597 correspondences between adjacent frames.
CoTracker3~\cite{karaev2025cotracker3} proposes the initial tracks, which annotators refine frame by frame in a custom tool, removing points that cannot be tracked (Fig.~\ref{fig:colon_tracking}).

Tab.~\ref{tab:pose_bench} reports the results on our annotated correspondences, in terms of the median Sampson distance and Pass@2, the fraction of correspondences with a distance of at most 2\,px.
Contrary to our expectation, endoscopy-specific estimators do not outperform general-purpose methods.
MegaSaM~\cite{li2025megasam} achieves the lowest error among methods that estimate poses for all frames, so we use it to annotate camera poses for all training videos.
This ranking also holds at longer frame gaps, where the rigidity assumption becomes weaker (Appendix~\ref{sec:suppl_pose}).

\begin{table}[t]
    \centering
    \caption{Camera pose accuracy on human-verified point tracks between adjacent frames. Sampson distance (Eq.~\ref{eq:sampson}) in pixels. Pass@2 counts correspondences without an estimated pose as failures.}
    \label{tab:pose_bench}
    \scriptsize
    \begin{tabular*}{0.8\columnwidth}{@{\extracolsep{\fill}}l cc@{}}
        \toprule
        Method & Median (px)$\downarrow$ & Pass@2$\uparrow$ \\
        \midrule
        \multicolumn{3}{@{}l}{\textit{Endoscopy-specific}} \\
        \quad Endo-SfMLearner~\cite{ozyoruk2021endoslam-endosfmlearner} & 0.78 & 0.79 \\
        \quad AF-SfMLearner~\cite{shao2022afsfmlearner} & 0.37 & 0.92 \\
        \quad EndoDAC~\cite{cui2024endodac}  & 0.36 & \textbf{0.93} \\
        \quad NR-SLAM~\cite{rodriguez2024nrslam} & 0.40 & 0.72 \\
        \quad OneSLAM~\cite{teufel2024oneslam} & 0.49 & 0.69 \\
        \midrule
        \multicolumn{3}{@{}l}{\textit{General-purpose}} \\
        \quad COLMAP~\cite{schonberger2016colmap} & 0.27 & 0.72 \\
        \quad DROID-SLAM~\cite{teed2021droidslam} & 0.34 & 0.91 \\
        \quad MASt3R-SLAM~\cite{murai2025mast3rslam} & 0.39 & 0.82 \\
        \quad ViPE~\cite{huang2025vipe} & 0.50 & 0.84 \\
        \quad VGGT-$\Omega$~\cite{wang2026vggtomega} & 0.44 & 0.90 \\
        \quad \textbf{MegaSaM}~\cite{li2025megasam} & \textbf{0.26} & \textbf{0.93} \\
        \bottomrule
    \end{tabular*}
\end{table}

\subsection{Condition Annotations}
\label{sec:data_conditions}
For the anatomical segment condition, we use the per-frame colon segment labels released for REAL-Colon~\cite{biffi2024realcolon}. We consider six segments from the cecum to the rectum and treat frames labeled as ileum or other as unlabeled. Each window is assigned the segment that covers more than half of its frames. The resulting 734 training windows comprise 347 transverse, 126 sigmoid, 105 descending, 92 ascending, 38 rectum, and 25 cecum windows, along with one unlabeled window.

REAL-Colon also provides bounding boxes for 132 polyps across 46 videos, covering 342,109 frames. We convert these boxes into segmentation masks using SAM2~\cite{ravi2025sam2}, with each box as a prompt. As a result, 360 training windows contain polyp masks, covering 92 polyps. For tool use and irrigation, we cut separate windows from the training procedures, which yields 92 training windows with instruments such as snares, forceps, or catheters and 29 training windows with irrigation. Appendix~\ref{sec:suppl_cond} details the condition windows and annotations.

\section{ColonoWorld}
\label{sec:method}

\begin{figure*}[t]
    \centering
    \includegraphics[width=\linewidth]{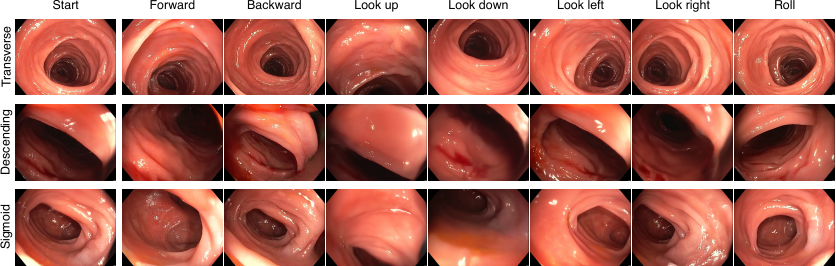}
    \caption{Generation under different camera motions. From the same initial image (Start), \methodname{} moves the camera forward and backward, rotates it up, down, left, and right, and rolls it as specified.}
    \label{fig:direction}
\end{figure*}

\methodname{} adapts an existing image-to-video generation model~\cite{agarwal2026cosmos3} to colonoscopy with a LoRA~\cite{hu2022lora}, keeping its base weights frozen.
It takes an initial image, a camera trajectory, and a colon tube (Fig.~\ref{fig:arch}), along with optional masks of lesions and tools and per-frame flags such as narrow-band imaging.
The tube is rendered along the camera trajectory, and the VAE of the video model encodes the rendering and the initial image into latents.
The trajectory is converted into per-frame ray maps, which a camera encoder injects into the transformer.

\subsection{Tube and Camera Conditions}
\label{sec:method_cond}
Unlike general videos, colonoscopy videos are captured through a fisheye lens inside a narrow, dark tube that follows the global structure of the colon.
We provide this structure to the model as a colon tube, a coarse 3D model of the colon around the camera.
The tube plays two roles in generation. First, its renderings provide a shared coarse geometric reference along the camera path.
In full-procedure generation, a whole-colon tube provides the global structure from the cecum to the rectum.
Second, it helps the model follow the camera.
A camera trajectory alone does not specify where the wall lies, and the tube improves how closely the generated video follows the camera (\cref{sec:exp_ablation}).
Compared with a detailed 3D model such as a hand-sculpted phantom~\cite{bobrow2023c3vd,golhar2026c3vdv2}, which is costly to build for many shapes, the tube describes only the coarse shape and leaves the appearance and fine details to the video model.

\paragraph{Colon tube construction.}
\label{sec:method_tube}
We model the tube around a smooth centerline.
Each cross-section is an ellipse, and its radius, aspect ratio, and orientation vary along the tube.
We also add haustral folds to the wall.
During training, we fit a tube to each window so that the dark lumen in its rendering matches the dark lumen in the real frames.
At test time, we build a tube from the initial image and the commanded trajectory, and for a full procedure we use a tube of the whole colon.
To evaluate full procedures, we build six whole-colon tubes with different lengths and colon courses.
Appendix~\ref{sec:suppl_tube} details the tube construction and shows the six tubes.

\begin{table*}[t]
    \centering
    \caption{Camera controllability and video generation quality.
    Best in \textbf{bold}, second best \underline{underlined}.
    $^\dagger$Implementation modified for our task and retrained on our training data. Appendix~\ref{sec:suppl_baselines} details the implementation of all baselines.}
    \label{tab:main}
    \footnotesize
    \setlength{\tabcolsep}{0pt}
    \newcolumntype{C}{>{\centering\arraybackslash}p{1.45cm}}
    \begin{tabular}{@{}l<{\hspace{28pt}}@{}CCCCCC@{}}
        \toprule
        \multirow{2}{*}[-0.5ex]{Method} & \multicolumn{3}{c}{Camera controllability} & \multicolumn{3}{c}{Video generation quality} \\
        \cmidrule(lr{6pt}){2-4} \cmidrule(l{6pt}r){5-7}
        & RotErr\rlap{$\downarrow$} & TransErr\rlap{$\downarrow$} & CamMC\rlap{$\downarrow$} & FVD\rlap{$\downarrow$} & FID\rlap{$\downarrow$} & IS\rlap{$\uparrow$} \\
        \midrule\addlinespace[2pt]
        \multicolumn{7}{@{}l}{\textit{Camera-controlled video generation}} \\
        Cosmos3-Nano 9D~\cite{agarwal2026cosmos3} & \phantom{0}6.09 & 10.30 & 13.92 & 1155 & \phantom{0}83.3 & 2.11 \\
        Wan2.2-Fun-5B-Control-Camera~\cite{wan2025wan} & \phantom{0}\underline{4.58} & 10.46 & 12.80 & 1424 & 113.9 & 2.28 \\
        CamI2V~\cite{zheng2024cami2v} & \phantom{0}4.99 & 11.87 & 14.34 & 3213 & 173.9 & \textbf{3.01} \\
        LingBot-World Base-Cam~\cite{team2026advancing} & \phantom{0}7.32 & 11.39 & 15.95 & 1145 & \phantom{0}80.6 & 2.14 \\
        UCPE$^\dagger$~\cite{zhang2026unified} & \phantom{0}4.73 & \phantom{0}\underline{6.97} & \underline{10.38} & \phantom{0}\underline{695} & \phantom{0}\underline{59.3} & 2.03 \\
        \midrule\addlinespace[2pt]
        \multicolumn{7}{@{}l}{\textit{Colonoscopy video generation}} \\
        Endora$^\dagger$~\cite{li2024endora} & 11.79 & 11.98 & 20.51 & 1738 & 113.6 & 1.92 \\
        FEAT$^\dagger$~\cite{wang2025feat} & 10.55 & 12.99 & 20.07 & 1583 & 119.4 & 1.53 \\
        MedSora$^\dagger$~\cite{wang2024optical_medsora} & 11.58 & 11.83 & 20.10 & 1445 & 128.9 & 1.86 \\
        EndoGen$^\dagger$~\cite{liu2025endogen} & \phantom{0}8.42 & 12.48 & 17.85 & 1982 & 145.0 & 2.31 \\
        SurgSora$^\dagger$~\cite{chen2025surgsora} & 11.19 & 10.59 & 18.97 & 2658 & 198.5 & \underline{2.34} \\
        \midrule
        \rowcolor{oursrow}
        Ours & \phantom{0}\textbf{3.20} & \phantom{0}\textbf{5.04} & \phantom{0}\textbf{7.09} & \phantom{0}\textbf{494} & \phantom{0}\textbf{49.7} & 2.32 \\
        \bottomrule
    \end{tabular}
\end{table*}

\paragraph{Conditioning on the tube and camera.}
We render the tube $\mathcal{T}$ along the camera poses $C_{1:N}$ with the fisheye camera model of each clip and a light at the camera, and encode the rendering with the frozen VAE $\mathcal{E}$ of the video model.
The resulting tube tokens are appended to the video tokens $z$~\cite{ju2025fulldit}:
\begin{equation}
    z^{\mathcal{T}} = \mathcal{E}\big(\mathrm{Render}(\mathcal{T}, C_{1:N})\big), \qquad \tilde{z} = \big[\, z \,;\, z^{\mathcal{T}} \,\big].
    \label{eq:tube_tokens}
\end{equation}
Each tube token shares the rotary position of the video token at the same frame and location, so every attention layer can compare each generated frame with the tube rendered from the same view.

For the camera, a camera encoder maps per-frame Pl\"ucker ray maps~\cite{he2025cameractrl}, computed with the fisheye lens model, to features that are added to the video tokens at the input and in every transformer block.
The initial image enters through the first-frame conditioning of the base model.
We first train the camera encoder with a LoRA, then freeze the encoder and train the LoRA together with the condition modules.
As shown in \cref{fig:direction}, \methodname{} generates videos that closely follow the input camera motion.

For a full procedure, we generate the video chunk by chunk.
Each chunk starts from the last generated frame of the previous one and is conditioned on the part of the tube along its path and on its colon segment.
Appendix~\ref{sec:suppl_train} gives the full training details.

\subsection{Lesion and Tool Conditions}
\label{sec:method_control}
Colonoscopy aims to find polyps, so we let users add polyps at user-specified locations on the colon, along with instruments, irrigation, and narrow-band imaging.
For a polyp placed on the tube wall, we represent its position, size, and shape as a mask video aligned with the tube rendering.
Instruments and irrigation are drawn into this mask video, and narrow-band imaging is given as a per-frame flag.
Like the tube, the mask video is encoded by the VAE and appended to the video tokens.
For training, we place each annotated polyp on the tube at its observed location, render its mask along the camera trajectory, and align the rendering with the segmentation mask.
Appendix~\ref{sec:suppl_condinput} details the condition inputs.

\section{Experiments}
\label{sec:experiments}

\subsection{\ph{Experimental setup}}
\paragraph{\ph{Datasets and inputs}}
We train and evaluate on REAL-Colon~\cite{biffi2024realcolon}, with training, validation, and test procedures split by patient.
The test set consists of 22 non-overlapping clips of 189 frames from four held-out patients, covering all six colon segments.
Each method receives the real first frame of a clip and its camera trajectory.
Since clinical videos come without a commanded trajectory, we use the trajectory recovered from the real clip, so that the real clip serves as the reference.
The trajectory is expressed relative to the first frame, in units of the median depth of that frame.
\methodname{} additionally receives the tube rendered along this trajectory.
We build this tube from the first frame, the trajectory, and the segment label, without reading any later frame.

\paragraph{Implementation details}
We fine-tune Cosmos3-Nano~\cite{agarwal2026cosmos3} at 1104$\times$832 and 24\,fps on windows of 65 or 189 frames, with a LoRA of rank 16 and a batch size of 8, using AdamW with a learning rate of $2\times10^{-5}$ for the LoRA and $10^{-4}$ for the tube and mask branches.
We first train the camera encoder together with the LoRA and freeze it, and then train the tube and mask branches with the LoRA.
Training takes about 1.4k B200 GPU-hours, or about 7 days on 8 GPUs.
At inference, we generate 189 frames with 35 sampling steps.
Appendix~\ref{sec:suppl_train} provides further training details.

\paragraph{\ph{Baselines and metrics}}
We compare with camera-controlled video generators, Cosmos3-Nano~\cite{agarwal2026cosmos3}, Wan2.2-Fun-5B-Control-Camera~\cite{wan2025wan}, CamI2V~\cite{zheng2024cami2v}, LingBot-World~\cite{team2026advancing}, and UCPE~\cite{zhang2026unified}, and with endoscopic video generators, Endora~\cite{li2024endora}, FEAT~\cite{wang2025feat}, MedSora~\cite{wang2024optical_medsora}, EndoGen~\cite{liu2025endogen}, and SurgSora~\cite{chen2025surgsora}.
None of the endoscopic generators takes a camera trajectory, and most of them do not take a first frame.
We therefore add the missing first-frame and camera conditioning to them and retrain them on our base training windows.
We also adapt UCPE, a recent camera control method with strong performance, to take a first frame and fine-tune it on the same data.

We evaluate camera controllability by recovering the camera trajectory of each generated video with MegaSaM~\cite{li2025megasam} and computing RotErr, TransErr~\cite{he2025cameractrl}, and CamMC~\cite{wang2024motionctrl} against the commanded trajectory, following the protocol of CamI2V~\cite{zheng2024cami2v}.
For the real-clip benchmark, both the reference and generated trajectories are reconstructed with MegaSaM. These scores therefore measure trajectory agreement within a common pose-estimation pipeline, which applies equally to all methods.

We measure video quality with FVD~\cite{unterthiner2018towards} and FID~\cite{heusel2017gans} against the real test videos, and with IS~\cite{salimans2016improved}.
Appendix~\ref{sec:suppl_baselines} details the implementation of all baselines.

\subsection{Camera Controllability and Video Quality}

\cref{tab:main} evaluates how closely each model follows the camera in short colonoscopy clips and how realistic the generated videos are.
\methodname{} achieves the best camera controllability, reducing RotErr, TransErr, and CamMC by 32\%, 28\%, and 32\% compared with UCPE, the strongest fine-tuned baseline.
Among the camera-controlled models without fine-tuning, Wan2.2-Fun attains the second-lowest RotErr, but all of them show high TransErr and CamMC, and their FVD is more than twice that of \methodname{}.
The colonoscopy models struggle to follow the camera even when trained with a camera condition on the same data, whereas \methodname{} reduces their CamMC by more than 60\%.

For video quality, \methodname{} reduces FVD by 29\% and FID by 16\% compared with UCPE, the best baseline on both metrics.
CamI2V attains the highest IS but the worst FVD, as IS does not compare generated frames with real videos.

\subsection{Full-Procedure Generation}
\label{sec:exp_fullproc}
A colonoscopy covers the whole colon in one continuous pass from the cecum to the rectum.
A colonoscopy world model should thus be able to generate a full procedure.
We test this on six whole-colon tubes of different shapes, each with a commanded path from the cecum to the rectum (Appendix~\ref{sec:suppl_tube}).
All methods start from the same real first frame and follow the same path.

\cref{tab:fullproc} reports camera controllability and video quality, averaged over the six shapes.
Since no real video corresponds to the synthetic colon shapes and commanded paths used here, we use real clips from held-out procedures as the reference.
To measure the general visual quality, we compute FVD and FID of each generated video against real clips randomly drawn from the same colon segments.
We repeat this random draw 10 times and report the mean.
\methodname{} reduces RotErr, TransErr, and CamMC by 41\%, 35\%, and 47\% over the best baseline on each metric, and achieves the best FVD and FID.

\begin{figure}[t]
    \centering
    \includegraphics[width=\linewidth]{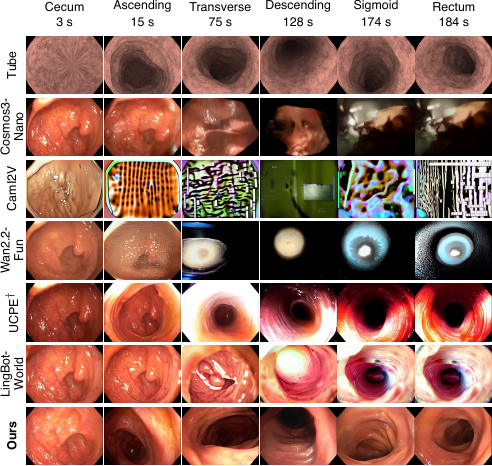}
    \caption{Full-procedure generation.
    Frames at the midpoint of each segment along the commanded cecum-to-rectum path, taken at the same frame index for every method. $^\dagger$Implementation modified for our task and retrained on our training data.}
    \label{fig:fullproc}
\end{figure}

\cref{fig:fullproc} shows a frame from each colon segment.
Cosmos3-Nano, CamI2V, and Wan2.2-Fun degrade severely and nearly lose the scene, and LingBot-World and UCPE also lose quality.
\methodname{} maintains its quality over the long generation and follows both the camera and the tube.

\begin{table}[t]
    \centering
    \caption{Full-procedure generation on six colon shapes (183 chunks).
    Values are averaged over the shapes, and a chunk whose camera cannot be re-estimated is scored with identity poses.
    Best in \textbf{bold}, second best \underline{underlined}. $^\dagger$Retrained on our data.}
    \label{tab:fullproc}
    \footnotesize
    \setlength{\tabcolsep}{2.5pt}
    \begin{tabular}{@{}l ccc cc@{}}
        \toprule
        & \multicolumn{3}{c}{Camera control} & \multicolumn{2}{c}{Video quality} \\
        \cmidrule(lr){2-4} \cmidrule(lr){5-6}
        Method & RotErr$\downarrow$ & TransErr$\downarrow$ & CamMC$\downarrow$ & FVD$\downarrow$ & FID$\downarrow$ \\
        \midrule
        Cosmos3-Nano~\cite{agarwal2026cosmos3}    & 7.03 & 9.33 & 14.20 & 2368 & 268.5 \\
        CamI2V~\cite{zheng2024cami2v}             & 5.64 & 10.76 & 13.82 & 4221 & 283.2 \\
        Wan2.2-Fun~\cite{wan2025wan}             & \underline{5.39} & 10.44 & 13.41 & 3424 & 257.6 \\
        UCPE$^\dagger$~\cite{zhang2026unified}    & 6.51 & \underline{5.45} & \underline{10.88} & 3147 & 250.9 \\
        LingBot-World~\cite{team2026advancing}    & 6.74 & 11.15 & 15.33 & \underline{2344} & \underline{210.3} \\
        \midrule
        Ours                                      & \textbf{3.18} & \textbf{3.52} & \textbf{5.81} & \textbf{1277} & \textbf{115.1} \\
        \midrule
        Real videos                               & -- & -- & -- & 333 & 34.3 \\
        \bottomrule
    \end{tabular}
\end{table}

\subsection{Lesion and Tool Control}
\label{sec:exp_control}
\begin{figure}[t]
    \centering
    \includegraphics{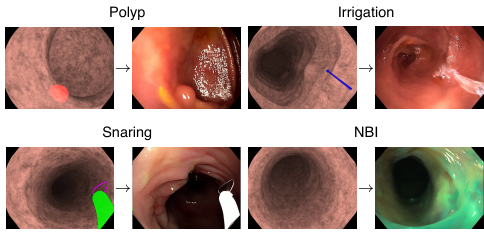}
    \caption{Lesion and tool control. Each pair shows the tube with the condition overlaid and the frame generated by \methodname{}.}
    \label{fig:control}
\end{figure}

\cref{fig:control} shows each condition overlaid on the tube rendering on the left and the generated frame on the right.
Polyps are given as red masks, irrigation as a blue water ribbon, and snaring as the snare loop and its sheath.
Narrow-band imaging is given as a per-frame flag, so its tube rendering carries no overlay.
In all cases, \methodname{} generates plausible lesions, instruments, and imaging modes at the specified locations.
The polyp takes the position and size of its mask and shows mucosal texture and specular highlights consistent with the surrounding wall.
Irrigation appears as a water jet along the given ribbon, and the snare and its sheath follow the drawn outline.
With the narrow-band imaging flag, the colors shift to the narrow-band palette while the lumen and folds still follow the tube.
Since the polyp mask is rendered on the tube along the camera trajectory, the generated polyp stays attached to the colon wall as the camera moves, while the instruments and irrigation are drawn relative to the scope and move with it.

\subsection{\ph{Ablation study}}
\label{sec:exp_ablation}
\paragraph{Tube and camera conditions.}
\cref{tab:ablation_cond} shows the effect of removing the tube and camera conditions in training and at test time.
Rows (a) and Ours$_{1000}$ compare training without and with the tube, each for 1,000 updates.
Adding the tube reduces RotErr, TransErr, and CamMC by 40\%, 50\%, and 45\% and also improves FVD and FID, showing that the tube is key to following the camera and generating realistic colonoscopy videos.

Rows (b), (c), and Ours remove the tube or the camera condition from our final model at test time.
Without the tube, camera controllability drops sharply, as CamMC more than doubles, and video quality also degrades.
Without the camera condition, video quality is nearly unchanged but all camera errors increase, indicating that the camera condition complements the tube in following the camera.

To measure whether the generated colon follows the tube, we compare the lumen of each generated frame with that of the tube rendering of the same frame.
We take the darkest 10\% of pixels in each image as the lumen and compute the distance between their centroids, normalized by the image diagonal.
Training with the tube lowers this lumen distance from 0.222 to 0.136, and removing the tube at test time raises it from 0.158 to 0.231, with an increase in 19 of the 22 clips.
Removing the camera condition changes it only slightly, from 0.158 to 0.168, so the lumen of the generated video is placed mainly by the tube.
We provide further details of the lumen distance in Appendix~\ref{sec:suppl_lumen}.

\paragraph{Full-procedure generation.}
\cref{tab:ablation_fullproc} shows the same trends as on the test set, and the effect of the tube is even stronger over a full procedure.
Training with the tube reduces RotErr, TransErr, and CamMC by 45\%, 56\%, and 52\%, compared with 40\%, 50\%, and 45\% on the test set, and lowers the lumen distance from 0.223 to 0.069.
Removing the tube at test time increases CamMC by 2.5 times, compared with 2.1 times on the test set, and raises the lumen distance from 0.106 to 0.246.
Without the camera condition, camera errors barely change but FVD and FID rise from 1277 to 1396 and from 115.1 to 127.2, indicating that the camera condition also helps keep the video realistic over a long procedure.

\begin{table}[t]
    \centering
    \caption{Ablation on the tube and camera conditions on the test set. Values are the mean and standard deviation over three seeds.}
    \label{tab:ablation_cond}
    \footnotesize
    \setlength{\tabcolsep}{1pt}
    \resizebox{\linewidth}{!}{%
    \begin{tabular}{@{}lccc ccccc c@{}}
        \toprule
        & Train & \multicolumn{2}{c}{Test} & \multicolumn{3}{c}{Camera control} & \multicolumn{2}{c}{Quality} & Lumen \\
        \cmidrule(lr){2-2} \cmidrule(lr){3-4} \cmidrule(lr){5-7} \cmidrule(lr){8-9} \cmidrule(l){10-10}
        & Tube & Tube & Cam. & RotErr$\downarrow$ & TransErr$\downarrow$ & CamMC$\downarrow$ & FVD$\downarrow$ & FID$\downarrow$ & Dist.$\downarrow$ \\
        \midrule
        (a) &  &  & \checkmark & 6.52$_{\pm.52}$ & 11.55$_{\pm.49}$ & 15.39$_{\pm.52}$ & 659$_{\pm163}$ & 65.5$_{\pm5.5}$ & 0.222 \\
        Ours$_{1000}$ & \checkmark & \checkmark & \checkmark & 3.94$_{\pm.23}$ & 5.76$_{\pm.99}$ & 8.41$_{\pm.91}$ & 541$_{\pm89}$ & 57.4$_{\pm2.3}$ & 0.136 \\
        \midrule
        (b) & \checkmark &  & \checkmark & 6.70$_{\pm.27}$ & 11.27$_{\pm.74}$ & 15.29$_{\pm.55}$ & 599$_{\pm136}$ & 60.1$_{\pm2.2}$ & 0.231 \\
        (c) & \checkmark & \checkmark &  & 3.62$_{\pm.57}$ & 5.42$_{\pm.46}$ & 7.77$_{\pm.82}$ & 480$_{\pm15}$ & 50.0$_{\pm2.2}$ & 0.168 \\
        Ours & \checkmark & \checkmark & \checkmark & 3.30$_{\pm.13}$ & 5.12$_{\pm.17}$ & 7.26$_{\pm.26}$ & 490$_{\pm24}$ & 49.4$_{\pm1.2}$ & 0.158 \\
        \bottomrule
    \end{tabular}}
\end{table}

\begin{table}[t]
    \centering
    \caption{Ablation on the tube and camera conditions on the full-procedure benchmark.}
    \label{tab:ablation_fullproc}
    \footnotesize
    \setlength{\tabcolsep}{1pt}
    \resizebox{\linewidth}{!}{%
    \begin{tabular}{@{}lccc ccccc c@{}}
        \toprule
        & Train & \multicolumn{2}{c}{Test} & \multicolumn{3}{c}{Camera control} & \multicolumn{2}{c}{Quality} & Lumen \\
        \cmidrule(lr){2-2} \cmidrule(lr){3-4} \cmidrule(lr){5-7} \cmidrule(lr){8-9} \cmidrule(l){10-10}
        & Tube & Tube & Cam. & RotErr$\downarrow$ & TransErr$\downarrow$ & CamMC$\downarrow$ & FVD$\downarrow$ & FID$\downarrow$ & Dist.$\downarrow$ \\
        \midrule
        (a) &  &  & \checkmark & 6.79 & 9.47 & 14.13 & 1392 & 160.9 & 0.223 \\
        Ours$_{1000}$ & \checkmark & \checkmark & \checkmark & 3.72 & 4.13 & 6.79 & 1239 & 143.4 & 0.069 \\
        \midrule
        (b) & \checkmark &  & \checkmark & 6.92 & 9.75 & 14.43 & 1425 & 143.6 & 0.246 \\
        (c) & \checkmark & \checkmark &  & 3.18 & 3.63 & 5.88 & 1396 & 127.2 & 0.127 \\
        Ours & \checkmark & \checkmark & \checkmark & 3.18 & 3.52 & 5.81 & 1277 & 115.1 & 0.106 \\
        \bottomrule
    \end{tabular}}
\end{table}

\section{Limitations}
\label{sec:limitations}

Since we learn from in vivo videos, we cannot measure how the colon actually moves, so we assume that the tube keeps its overall shape within each window.
Strong deformation, such as a peristaltic wave, could in principle be expressed by changing the tube over time.
We do not consider it in this work, as our training tubes do not include such waves and the actual deformation they cause is unknown.
When we generate a full procedure chunk by chunk, the tube provides a shared coarse geometric reference, but fine details such as the mucosal texture and small vessels may differ when the camera returns to the same region.
Finally, we have not yet trained a robotic policy that navigates or intervenes inside \methodname{}, nor measured whether such a policy improves the success rate of colonoscopy, which remains future work.

\section{Conclusion}
\label{sec:conclusion}
We present \methodname{}, a colonoscopy world model that generates video along a commanded camera trajectory, guided by a coarse, editable tube of the colon.
Trained on in vivo videos with recovered camera poses, it generates a full procedure from the cecum to the rectum and places polyps, instruments, and irrigation where and when they are specified.
It reduces camera trajectory error by 32\% and FVD by 29\% relative to the strongest baseline.
Since every generated video comes with its camera trajectory, anatomy, and lesion locations, we hope that \methodname{} supports the development of camera tracking, navigation assistance, and lesion detection in colonoscopy.

{
    \small
    \bibliographystyle{ieeenat_fullname}
    \bibliography{main}
}

\clearpage
\onecolumn
\setcounter{page}{1}
{\renewcommand\twocolumn[1][]{#1}\maketitlesupplementary}
\appendix

\addtocontents{toc}{\protect\setcounter{tocdepth}{2}}
\begingroup
\makeatletter
\renewcommand*\l@section[2]{\ifnum\c@tocdepth>\z@
  \addpenalty{-\@highpenalty}\vskip 1.8em plus 1pt
  {\leftskip 0em \rightskip 2.5em plus 1fil \parfillskip -\rightskip
   \parindent 0em \@tempdima 2em
   \advance\leftskip\@tempdima \hskip -\leftskip
   \large #1\nobreak\leaders\hbox{$\m@th\mkern 6mu\hbox{.}\mkern 6mu$}\hfill
   \nobreak\hb@xt@ 2.5em{\hss\large #2}\par}\fi}
\renewcommand*\l@subsection[2]{\ifnum\c@tocdepth>\@ne
  \vskip 0.55em plus 1pt
  {\leftskip 2.3em \rightskip 2.5em plus 1fil \parfillskip -\rightskip
   \parindent 0em \@tempdima 2.2em
   \advance\leftskip\@tempdima \hskip -\@tempdima
   \normalsize #1\nobreak\leaders\hbox{$\m@th\mkern 6mu\hbox{.}\mkern 6mu$}\hfill
   \nobreak\hb@xt@ 2.5em{\hss\normalsize #2}\par}\fi}
\def\numberline#1\hskip -1em.{\hb@xt@\@tempdima{#1.\hfil}}
\renewcommand\contentsname{Contents}
\renewcommand\section{\@startsection{section}{1}{\z@}{0pt}{1.5em}{\centering\Large\bfseries}}
\makeatother
\hypersetup{linkcolor=black}
\vspace*{\stretch{1}}
\begin{center}
\begin{minipage}{0.95\linewidth}
\tableofcontents
\end{minipage}
\end{center}
\vspace*{\stretch{2}}
\endgroup
\clearpage
\twocolumn

\section{Dataset Details}
\label{sec:suppl_data}

\subsection{Base Windows}
\label{sec:suppl_base}

\paragraph{Source data.}
REAL-Colon~\cite{biffi2024realcolon} contains 60 colonoscopy procedures from four clinical studies across six centers, with 2.76M frames (28.1\,h).
We use the per-frame anatomical segment labels released with the dataset and keep only the withdrawal phase, which we define as the frames labeled with one of the seven segment classes from the ileum to the rectum (17.7\,h).

\paragraph{Filtering.}
We split the withdrawal footage into 6,309 segments of 10\,s and score each segment on frames sampled at about 3\,fps.
We reject segments that are mostly dark or overexposed, show no dark lumen region, or are blurry, which leaves 5,367 segments.
A segment counts as mostly dark or overexposed if more than 20\% of its frames have a mean gray level below 25 or more than half of their pixels above 245.
It shows no lumen if the median area of its largest dark region is below 2.9\% of the field of view, and it is blurry if the median variance of its Laplacian is below 991, which are the 5th and 10th percentiles over all segments.
We then keep segments whose median optical-flow magnitude lies between the 5th and 80th percentiles, which removes near-static views and fast scope motion and leaves 4,024 segments.
Finally, a linear probe on DINOv2 features detects whether the lumen is visible in each frame, and we cut clips of 4 to 9.5\,s in which the lumen is visible in at least 90\% of the frames.
This gives 1,017 clips from 57 procedures.

\paragraph{Procedures and split.}
We use a fixed fisheye (KB4) camera model for each scope group and keep the 41 procedures whose sensor and crop match one of these models and that yield at least one clip.
We split them by patient into 36 training, 1 validation, and 4 test patients.
We further drop the 8 training procedures of one center whose videos show on-screen text inside the image, since masking the text would also remove tissue.
This leaves 28 training procedures.

\paragraph{Training windows.}
Each training window has 65 or 189 consecutive frames.
The final set contains 734 windows (598 of 65 frames and 136 of 189 frames) cut from 453 clips, including 220 windows from additional passes, 37 over the cecum and rectum, 98 over the flexures, and 85 over the other segments.
The windows contain 44,824 unique frames (28.4\,min), whose segments are listed in Tab.~\ref{tab:suppl_segments}.
All frames are played at 24\,fps, so videos recorded at 25 and 29.97\,fps are slowed by 4\% and 20\%.

\paragraph{Content check.}
After pose estimation, we remove windows that still show instruments, irrigation, dye, or special-light imaging.
An ensemble of linear SVM classifiers on DINO, color, texture, and ResNet18 features flags these classes per frame, and a window is rejected if any class fires on at least 3\% of its frames.
Specular highlights, small droplets, blood, and residue do not lead to rejection on their own.

\paragraph{Test set.}
The test set contains 22 clips of 189 frames from the four test patients, none of whom appear in training.

\begin{table}[h]
    \centering
    \caption{Anatomical segments of the 44,824 unique frames in the training windows.}
    \label{tab:suppl_segments}
    \small
    \begin{tabular}{@{}lr@{}}
        \toprule
        Segment & Frames \\
        \midrule
        Transverse & 19,642 \\
        Sigmoid & 7,609 \\
        Descending & 7,128 \\
        Ascending & 6,155 \\
        Rectum & 2,189 \\
        Cecum & 2,036 \\
        Ileum & 65 \\
        \midrule
        Total & 44,824 \\
        \bottomrule
    \end{tabular}
\end{table}

\subsection{Condition Windows}
\label{sec:suppl_cond}
Besides the 734 base windows, the final model is trained on 395 condition windows (Tab.~\ref{tab:suppl_cond}).
Seven of these procedures come from the center excluded from the base windows and contribute only polyp and narrow-band imaging windows.
Condition windows may overlap base windows in time, and 28 of them include footage from the insertion phase.

\paragraph{Polyps.}
We convert the REAL-Colon boxes into masks with SAM2~\cite{ravi2025sam2}, place each polyp on the colon tube, and use its rendering, aligned with the masks, as the condition.
Polyp masks are available in 214 base windows and 146 additional windows, covering 92 polyps.

\paragraph{Instruments.}
Instrument masks are obtained with SAM2 from boxes and points of a tool detector or from annotated points, and propagated through each window.
In 33 snare windows, annotators also draw the snare wire as a polyline on keyframes.

\paragraph{Irrigation.}
Annotators mark where the water jet hits the wall on up to three frames per window, and CoTracker3~\cite{karaev2025cotracker3} tracks this point through the window.

\paragraph{Narrow-band imaging.}
The 128 narrow-band imaging windows carry per-frame labels of the light mode, and 40 of them switch between white light and narrow-band imaging.

\begin{table}[h]
    \centering
    \caption{Condition windows of the final model.}
    \label{tab:suppl_cond}
    \small
    \begin{tabular}{@{}lrr@{}}
        \toprule
        Condition & Windows & With insertion \\
        \midrule
        Polyp (additional) & 146 & 17 \\
        Instrument & 92 & 0 \\
        Irrigation & 29 & 0 \\
        Narrow-band imaging & 128 & 11 \\
        \midrule
        Total & 395 & 28 \\
        \bottomrule
    \end{tabular}
\end{table}

\section{Camera Pose Benchmark at Longer Frame Gaps}
\label{sec:suppl_pose}

\paragraph{Annotation.}
The annotation tool seeds query points, and CoTracker3~\cite{karaev2025cotracker3} proposes their tracks.
Annotators correct the positions and assign each point one of four states (visible, occluded, out of frame, or uncertain).
A frame range counts as confirmed only after an annotator has viewed every frame in it at the original speed, and a confirmation becomes invalid if the track is changed afterwards.
Within a confirmed range, a proposed point with a position counts as visible, and one without a position counts as out of frame.
We use the 11 fully reviewed clips from 10 patients, with 92 tracks over 2,036 frames, of which 13,708 track-frames are visible.
The annotated clips are taken from training patients, since the benchmark selects the pose estimator used to annotate the training data rather than evaluating the generator.

\paragraph{Metric.}
A correspondence between frames $i$ and $i{+}L$ is eligible if the point is confirmed visible in both.
For each eligible correspondence, we compute the Sampson distance (Eq.~\ref{eq:sampson}) from the estimated relative pose.
We report the median distance over all correspondences, pooled across clips.
We also report Pass@2, the fraction of eligible correspondences with a distance of at most 2\,px, where correspondences without an estimated pose count as failures.
The main paper uses $L{=}1$, and Tab.~\ref{tab:suppl_pose_lags} adds $L \in \{4, 16, 32\}$, for which the local rigidity assumption becomes weaker.

\paragraph{Implementation.}
All estimators run on the same clips, which we rectify from the KB4 model to $640{\times}480$ pinhole images with a shared intrinsic matrix per clip.
We use the released weights of each method without training or fine-tuning on our data.
Translation scale is not aligned, since the Sampson distance depends only on the direction of translation.
Frames for which a method fails or loses tracking are marked invalid rather than interpolated.
Tab.~\ref{tab:suppl_pose_impl} summarizes the settings of each method.

\paragraph{Results.}
Among the methods in Tab.~\ref{tab:suppl_pose_lags}, MegaSaM achieves the lowest median error at every frame gap and the highest or tied-highest Pass@2, tied with EndoDAC at $L{=}1$ and with DROID-SLAM at $L{=}32$.
Self-supervised endoscopy-specific estimators degrade most at longer gaps.

\begin{table*}[t]
    \centering
    \caption{Settings of the pose estimators. All methods use released weights, and $\mathbf{K}$ denotes the shared per-clip intrinsics.}
    \label{tab:suppl_pose_impl}
    \scriptsize
    \setlength{\tabcolsep}{4pt}
    \begin{tabular}{@{}l >{\raggedright\arraybackslash}p{3.8cm} >{\raggedright\arraybackslash}p{1.8cm} >{\raggedright\arraybackslash}p{3.0cm} >{\raggedright\arraybackslash}p{3.6cm}@{}}
        \toprule
        Method & Weights & Input & Intrinsics & Scale and depth \\
        \midrule
        Endo-SfMLearner & Released pose network & $832{\times}256$ & Not used by the network & Learned monocular, arbitrary scale \\
        AF-SfMLearner & Released pose encoder and decoder & $320{\times}256$ & Not used by the network & Learned monocular, arbitrary scale \\
        EndoDAC & Released pose and intrinsics heads & $320{\times}256$ & Predicted, $\mathbf{K}$ used for scoring & Learned monocular, arbitrary scale \\
        NR-SLAM & None (classical) & $640{\times}480$ & $\mathbf{K}$ & Median depth normalized \\
        OneSLAM & CoTracker (stride 4, window 8) & $640{\times}480$ & $\mathbf{K}$ & Constant initial depth \\
        \midrule
        DROID-SLAM & Released model & ${\approx}384{\times}512$ & $\mathbf{K}$, rescaled & Monocular RGB, no depth input \\
        MASt3R-SLAM & Released metric MASt3R and retrieval codebook & 512 & $\mathbf{K}$ (known-intrinsics mode) & Sim(3) tracking \\
        ViPE & MoGe-2 ViT-L and DROID-SLAM & $640{\times}480$ & $\mathbf{K}$, not optimized & MoGe-2 depth prior \\
        VGGT-$\Omega$ & Released 1B model & $640{\times}480$, 32-frame windows with 8-frame overlap & Predicted, $\mathbf{K}$ used for scoring & Feed-forward, windows chained by a rigid transform and the median overlap scale \\
        MegaSaM & Released model, Depth Anything and UniDepthV2 priors & Internal resizing & $\mathbf{K}$, not optimized & Monocular depth priors, bundle adjustment, no CVD refinement \\
        \bottomrule
    \end{tabular}
\end{table*}

\begin{table*}[t]
    \centering
    \caption{Camera pose accuracy on human-verified point tracks at frame gap $L$. Each cell shows the median Sampson distance in pixels ($\downarrow$) / Pass@2 ($\uparrow$). Best in \textbf{bold}.}
    \label{tab:suppl_pose_lags}
    \small
    \begin{tabular}{@{}l cccc@{}}
        \toprule
        Method & $L{=}1$ & $L{=}4$ & $L{=}16$ & $L{=}32$ \\
        \midrule
        \multicolumn{5}{@{}l}{\textit{Endoscopy-specific}} \\
        \quad Endo-SfMLearner~\cite{ozyoruk2021endoslam-endosfmlearner} & 0.78 / 0.79 & 2.70 / 0.41 & 8.07 / 0.15 & 12.58 / 0.10 \\
        \quad AF-SfMLearner~\cite{shao2022afsfmlearner} & 0.37 / 0.92 & 1.11 / 0.67 & 3.45 / 0.34 & 5.61 / 0.22 \\
        \quad EndoDAC~\cite{cui2024endodac} & 0.36 / \textbf{0.93} & 1.13 / 0.68 & 3.41 / 0.34 & 5.95 / 0.20 \\
        \quad NR-SLAM~\cite{rodriguez2024nrslam} & 0.40 / 0.72 & 1.02 / 0.55 & 2.58 / 0.36 & 4.20 / 0.24 \\
        \quad OneSLAM~\cite{teufel2024oneslam} & 0.49 / 0.69 & 0.87 / 0.65 & 2.19 / 0.48 & 3.48 / 0.38 \\
        \midrule
        \multicolumn{5}{@{}l}{\textit{General-purpose}} \\
        \quad DROID-SLAM~\cite{teed2021droidslam} & 0.34 / 0.91 & 0.68 / 0.78 & 1.53 / 0.58 & 2.28 / \textbf{0.47} \\
        \quad MASt3R-SLAM~\cite{murai2025mast3rslam} & 0.39 / 0.82 & 0.86 / 0.68 & 1.90 / 0.47 & 2.86 / 0.37 \\
        \quad ViPE~\cite{huang2025vipe} & 0.50 / 0.84 & 1.04 / 0.69 & 2.10 / 0.49 & 3.17 / 0.38 \\
        \quad VGGT-$\Omega$~\cite{wang2026vggtomega} & 0.44 / 0.90 & 1.05 / 0.69 & 2.64 / 0.42 & 4.07 / 0.30 \\
        \quad MegaSaM~\cite{li2025megasam} & \textbf{0.26} / \textbf{0.93} & \textbf{0.58} / \textbf{0.81} & \textbf{1.39} / \textbf{0.61} & \textbf{2.26} / \textbf{0.47} \\
        \bottomrule
    \end{tabular}
\end{table*}

\section{Colon Tube Construction}
\label{sec:suppl_tube}

\begin{figure*}[t]
    \centering
    \includegraphics[width=0.8\linewidth]{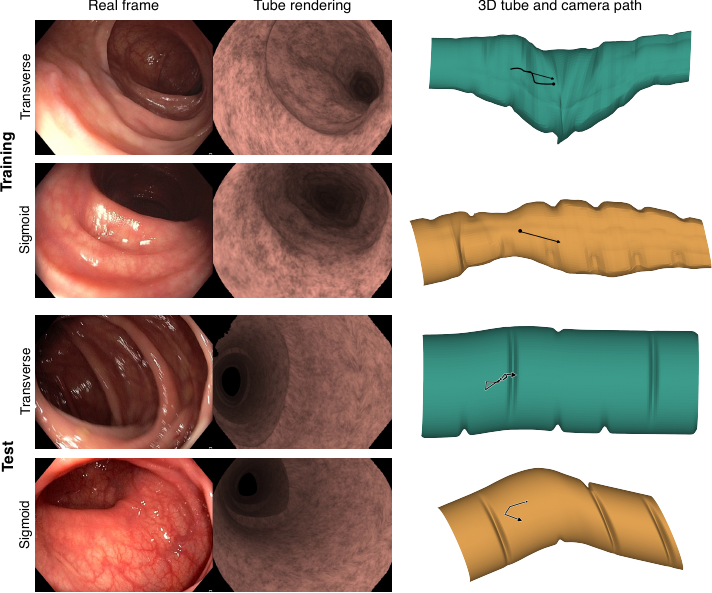}
    \caption{Examples of training and test tubes.}
    \label{fig:suppl_tubes}
\end{figure*}

Training, test, and full-procedure tubes share one representation and renderer (Fig.~\ref{fig:suppl_tubes}).
All lengths are in units of a scale $R$, and no tube is metric.

\paragraph{Representation.}
A tube is described by five profiles $q_0,\dots,q_4$ along its axis $z$, each given at 24 knots and interpolated with PCHIP.
The first two bend the centerline away from the axis $\mathbf{b}_2$, as $\mathbf{c}(z)=\mathbf{o}+z\,\mathbf{b}_2+q_0(z)\,\mathbf{b}_0+q_1(z)\,\mathbf{b}_1$.
The other three shape the cross-section, an ellipse with area-equivalent radius $r=e^{q_2}$, semi-axes $re^{\pm\epsilon}$ with $\epsilon=\|(q_3,q_4)\|$, and orientation $\tfrac{1}{2}\operatorname{atan2}(q_4,q_3)$.
Haustral folds narrow the tube at regular intervals by multiplying the elliptical radius by
\begin{equation}
F(s,\phi)=1-\max_{j}\max_{k\in\mathcal{K}_j} d\,g_{j}(s,\phi)\,A_k(\phi),
\label{eq:tube_fold_factor}
\end{equation}
where $s$ is the arc length and $\phi$ the azimuth.
The $j$-th fold is a Gaussian $g_j$ of width $w$ along $s$, the folds are spaced $\Delta$ apart, and $d$ is their relative depth.
Around the circumference, $A_k$ is a $\cos^2$ window over one of three sectors $120^\circ$ apart, and $\mathcal{K}_j$ holds the sectors covered by fold $j$.
Consecutive folds cover two, one, and one sectors in turn, so each fold forms a partial crescent like the semilunar folds of the colon.
A tube has 141 parameters, including six fold parameters, and is meshed with 72 azimuths per ring.

\paragraph{Rendering.}
Each pixel is lifted to a ray with the KB4 fisheye model of the clip, the same rays used for the Pl\"ucker condition.
Mitsuba~3 is used only for ray--mesh intersection.
The wall is lit by a single light at the camera.
For a ray $\mathbf{d}$ that hits the wall at distance $\ell$ with normal $\mathbf{n}$, the shading is
\begin{equation}
\begin{aligned}
\sigma={}&\min\!\left(\frac{1.25\,(0.2+0.8\,|\mathbf{n}^{\top}\mathbf{d}|)}{\max(\ell/R,0.08)^{2}}+0.012,1\right)\\
&\cdot e^{-0.055\max(\ell/R-2,0)} .
\end{aligned}
\label{eq:tube_shading}
\end{equation}
It decreases with the square of the distance and is darker where the ray grazes the wall.
The pixel value is $255\,\operatorname{clip}(a\sigma,0,1)^{1/2.2}$, where the albedo $a$ comes from one fixed mucosa texture mapped by arc length and circumference.

\begin{figure*}[p]
    \centering
    \includegraphics[width=\linewidth,height=0.86\textheight,keepaspectratio]{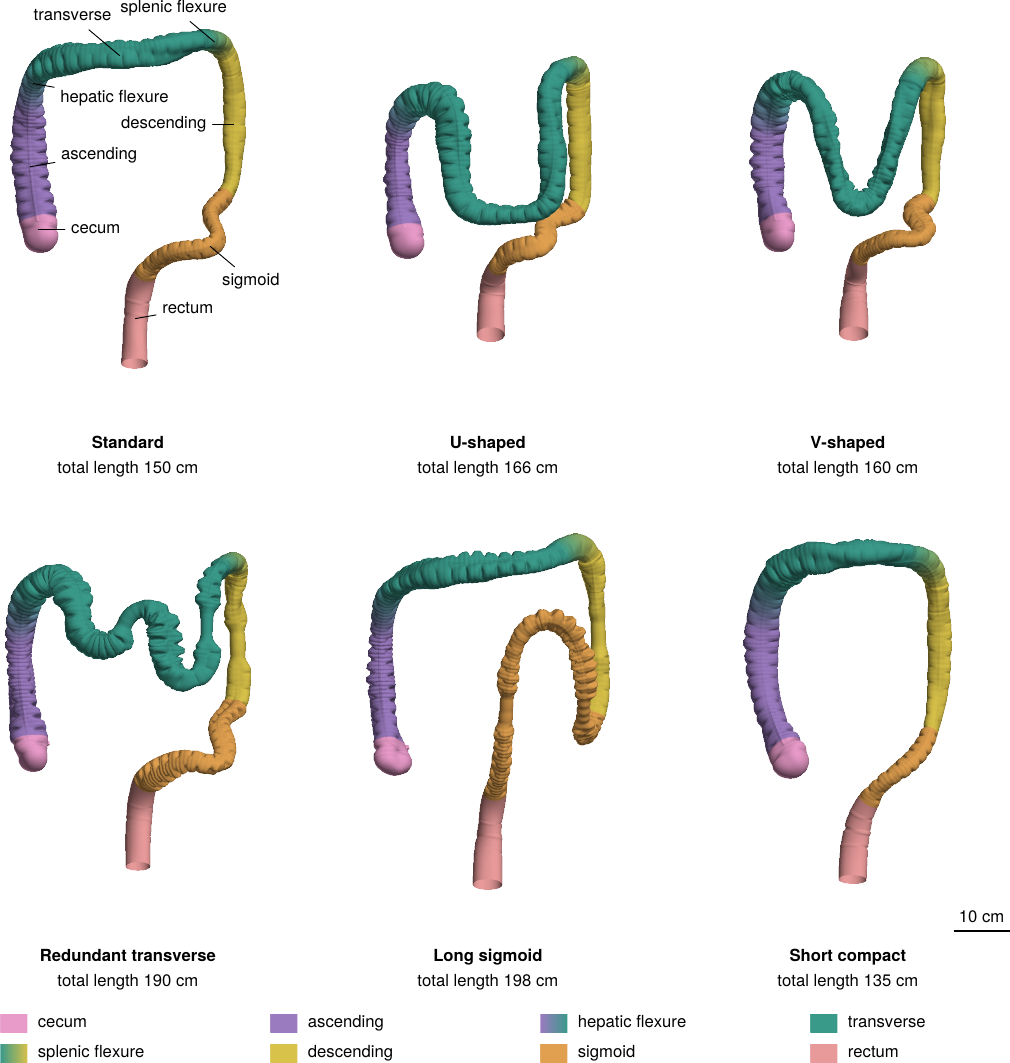}
    \caption{The six whole-colon tubes of the full-procedure benchmark, shown from the anterior view at a common scale.}
    \label{fig:suppl_fullcolon6}
\end{figure*}

\paragraph{Training tubes.}
Each training window gets its own tube, fitted to its MegaSaM poses~\cite{li2025megasam}, a static signed-distance field (SDF) of the wall, and its real frames.
We first fit an ellipse to the wall at each of 24 slices and the folds to the radial deficit of the wall.
We then extend the tube beyond the front-most camera with a 15-parameter neck, bend, and taper.
We fit 14 of these parameters by differential evolution and choose the remaining dilation factor from a fixed set, so that the rendered opening matches the real dark lumen in presence, position, and size.
Every tube keeps all cameras inside with a clearance of at least $0.3R$ and extends at least $7R$ past the neck, and tubes of cecum windows end in a closed cap.

\paragraph{Test tubes.}
A test tube is built only from frame~0, the commanded path, the camera model and the colon-segment label, with no optimization.
Translations are expressed in units of $R$, which we set to 0.3 times the median depth of frame~0.
The centerline follows the smoothed camera path along its principal direction, from $1.75R$ behind the rear-most camera to $8.8R$ beyond the front-most one.
It is shifted and bent toward the lumen detected in frame~0.
The radius is the median of training tubes in the same segment, and the folds are drawn from the reliable fold fits of the training tubes, with a random seed derived from the clip identifier.
The radius is then scaled so that every camera keeps a margin of $0.36R$.

\paragraph{Full-procedure tubes.}
For the full-procedure benchmark, we build six whole-colon tubes that cover common variants of the colon course (Fig.~\ref{fig:suppl_fullcolon6}).
Each tube follows per-segment centerline control points from the rectum to the cecum and keeps the normal order of the segments.
The shapes differ in the length and mobility of the segments, the height and angle of the flexures, and the caliber of the segments.
The standard shape has a nearly horizontal transverse colon.
The U-shaped, V-shaped, and redundant-transverse shapes vary the course of the transverse colon, the long-sigmoid shape adds a 72\,cm sigmoid loop, and the short compact shape has a high transverse colon and a mobile cecum.
Their nominal lengths range from 135 to 198\,cm.
The walls include haustral and semilunar folds, the rectal valves, and a cecal pouch.
The redundant-transverse and long-sigmoid shapes add deeper folds and contraction rings, and the redundant-transverse shape also has two flattened stretches of the transverse colon.

The camera dwells in the cecum for the first chunk and then withdraws to the rectum.
Its step per frame is scaled with the local tube radius so that the withdrawal speed matches the training data in each segment, and lateral and rotational offsets are replayed from training windows while keeping a clearance of $0.3R$ from the wall.
Each path is cut into 25 to 36 chunks of 189 frames that share one boundary frame, which gives 183 chunks over the six shapes.
The renderer is the same as for the training tubes, except that the texture is projected from world coordinates.

\section{Training Details}
\label{sec:suppl_train}

\paragraph{Conditioning.}
We render the tube at the resolution and frame rate of the video.
The tube tokens enter the model through a zero-initialized linear projection, so the model initially behaves as the base model.
An embedding of the colon-segment label is added to the tube tokens.
Camera poses are expressed relative to the first frame, and the ray maps are subsampled in time to match the temporal compression of the VAE.
All clips share the same generic text prompt.

\paragraph{Optimization.}
We add LoRA adapters of rank 16 and scale 32 to the query, key, value, and output projections of every attention layer.
We train with the rectified flow objective with a timestep shift of 10 and a batch size of 8, using AdamW with $\beta=(0.9, 0.999)$, no weight decay, gradient clipping at 1.0, and an EMA of 0.99.
After a short linear warm-up, the learning rate is kept constant.
We first train the camera branch together with the LoRA adapters at a learning rate of $10^{-4}$ for 1{,}200 updates and then freeze the camera branch.
We then train the LoRA adapters at a learning rate of $2\times10^{-5}$ and the tube token module at $10^{-4}$, first with the tube alone for 15{,}500 updates and then with all conditions for 2{,}500 updates.
In the second part, the mask and flag modules are trained at $10^{-4}$, except the mask input projection and, for the last 1{,}500 updates, the flag adapter, which use $10^{-3}$.
The full training takes 19{,}200 updates on windows of 65 or 189 frames at 24\,fps.

\paragraph{Dropout and inference.}
During training, we replace the tube with a null tube for 10\% of the samples and drop both the tube and the camera for 5\%.
To make the model robust to imperfect hand-off frames in full-procedure generation, we also corrupt the first conditioning frame in pixel or latent space for about half of the samples.
At inference, we use 35 sampling steps and a guidance scale of 1 for the tube and the text.
For a full procedure, each chunk of 189 frames starts from the last generated frame of the previous chunk without added noise, so consecutive chunks share one frame, and we apply no drift correction.

\section{Lesion and Tool Conditions}
\label{sec:suppl_condinput}

\paragraph{Mask video.}
All lesion and tool conditions are drawn into one RGB mask video, with polyps in red, instruments in green, water in blue, and the snare wire in magenta.
The classes are alpha-blended in the order polyp, water, instrument, and snare, with a soft edge of 3 pixels.
The snare wire is drawn as line segments of 8 pixels in width through the annotated points, with a disc at its tip.
Irrigation is drawn as a ribbon from the nozzle to the tracked point where the water hits the wall.
A class whose mask is not supplied is treated as unknown, whereas a supplied empty mask states that the class does not appear in the window.

\paragraph{Mask tokens.}
The VAE encodes the mask video, and the resulting tokens are appended to the video tokens as a separate block that shares the rotary positions of the video tokens.
The mask tokens have their own zero-initialized projector and their own time, height, width, and type embeddings.

\paragraph{Per-frame flags.}
For each latent frame, 13 numbers encode the light mode and the presence of instruments and irrigation, whether each of these states is known, and which mask classes are supplied.
A zero-initialized linear layer maps these numbers to the token dimension, and the result is added to the video tokens.

\paragraph{Dropout and guidance.}
During training, we mark the mask of each class and the flags as unknown with a probability of 0.10 each, and all lesion and tool conditions together with a probability of 0.05.
At inference, we apply classifier-free guidance to the mask with a scale of 3, using the same tube without masks as the unconditional input.

\section{Baseline Implementation}
\label{sec:suppl_baselines}

\subsection{Common Protocol}
\paragraph{Inputs and outputs.}
Every baseline receives the real first frame at $1104{\times}832$, the KB4 fisheye calibration of the scope group~\cite{kannala2006generic}, and the commanded trajectory as 189 camera poses relative to the first frame. These are the inputs of \methodname{} except for the colon-segment label, which it uses to build its tube and adds to the tube tokens.
Cosmos3-Nano takes only the poses, since its action interface has no camera intrinsics.
No baseline receives depth maps, future frames, or any file used for scoring.
Models that accept text (Cosmos3-Nano, Wan2.2-Fun, CamI2V, LingBot-World, and UCPE$^\dagger$) receive the fixed prompt of \methodname{}, \textit{``Egocentric colonoscopy video showing the colon lumen and mucosal surface. The video is 7.9 seconds long and is of 24 FPS. This video is of 832x1104 resolution.''}
The released models generate at or near their native resolution, namely $1104{\times}832$ for Cosmos3-Nano (its 720p tier), $1120{\times}832$ for Wan2.2-Fun, $512{\times}320$ for CamI2V, and $720{\times}544$ for LingBot-World (its 480p tier).
The retrained models generate on canvases of $1120{\times}832$ for UCPE$^\dagger$, $288{\times}208$ for Endora$^\dagger$ and FEAT$^\dagger$, $256{\times}256$ for MedSora$^\dagger$ and SurgSora$^\dagger$, and $128^2$ per frame for EndoGen$^\dagger$.
We resize the first frame uniformly and pad it with black borders instead of cropping it, which keeps the full field of view, and map the output back to $1104{\times}832$, 189 frames, and 24\,fps.
LingBot-World generates 129 frames at its native 16\,fps along the trajectory resampled to 16\,fps, and we convert its output to 24\,fps by repeating the frame nearest in time.

\paragraph{Camera representation.}
For every model that takes rays or Pl\"ucker embeddings, the camera condition is computed from the calibrated KB4 rays of each pixel or patch.
Replacing the KB4 model with the closest pinhole camera would displace the rays by 83\,px on average and by 233\,px (median) in the outer 5\% of the field of view, giving the baselines a wrong camera geometry.
Ray directions outside the field of view and in the padding are set to zero. UCPE$^\dagger$ masks these patches instead, and SurgSora$^\dagger$ moves their flow outside the canvas.
Translations keep the unit of the trajectory, which is the median depth of the first frame, except for Cosmos3-Nano and LingBot-World as described below.

\paragraph{Long videos.}
Cosmos3-Nano, Wan2.2-Fun, LingBot-World, and UCPE$^\dagger$ generate all frames in one pass.
The other models generate chunks of their native length $L$, which is 16 for CamI2V, Endora$^\dagger$, FEAT$^\dagger$, MedSora$^\dagger$, and EndoGen$^\dagger$ and 25 for SurgSora$^\dagger$.
Each chunk starts from the last generated frame of the previous chunk, so the chunks start at frames $0, L{-}1, 2(L{-}1), \dots$, and the camera poses are expressed relative to that frame.
The last chunk is moved back to end at the last frame, so it starts inside the previous chunk and replaces its last 7 frames (4 frames for SurgSora$^\dagger$).
Endora$^\dagger$ and FEAT$^\dagger$ pass this frame on as a latent and EndoGen$^\dagger$ as tokens, without decoding.

\subsection{Baselines with Released Weights}
\label{sec:suppl_released}
Cosmos3-Nano~\cite{agarwal2026cosmos3} is used in its action-conditioned image-to-video mode.
Here the camera motion enters as a 9D action per frame relative to the previous frame, consisting of the translation and the first two columns of the relative rotation, with 188 actions for 189 frames.
Since the action space has no metric scale, we multiply the translations by a constant of 163.6 so that the median step per frame matches the official camera example.
We run it at 24\,fps with the guardrails disabled.
Wan2.2-Fun-5B-Control-Camera~\cite{wan2025wan} and CamI2V~\cite{zheng2024cami2v} take Pl\"ucker embeddings, which we compute from KB4 rays. For CamI2V, we also compute the epipolar attention mask from KB4 bearing vectors.
LingBot-World Base-Cam~\cite{team2026advancing} takes per-pixel ray directions and poses, for which we use KB4 rays.
It normalizes the translation magnitude internally, and we disable its on-screen control overlay.

\subsection{Retrained Baselines ($^\dagger$)}
\paragraph{Training protocol.}
All $^\dagger$ models are trained on the same 734 windows as \methodname{}. The 27 forward-moving windows are repeated four times, and the pool is shuffled in the same order for every model.
Validation and test patients are excluded by the data loader.
We train on consecutive frames without flips or crops, which would break the correspondence between the frames and the camera poses.
Each model is trained until it has seen 32{,}000 windows.
We keep hourly checkpoints and select the one with the lowest loss on the validation windows.
Every layer added for the first frame or the camera is zero-initialized.
Tab.~\ref{tab:suppl_retrain} lists the initialization and size of each model.

\paragraph{UCPE$^\dagger$.}
We port the camera-encoding code of UCPE~\cite{zhang2026unified}, without its weights, to Wan2.2-Fun-5B-Control-Camera and remove the backbone's own Pl\"ucker adapter so that the camera enters only through UCPE.
We train only the UCPE module with the backbone frozen. The module uses the relative ray with absolute orientation map, computed from KB4 rays on 32\,px patches.
Since training runs in the backbone's framework (VideoX-Fun), we use the optimizer settings of its example control-training script, with a learning rate of $2{\times}10^{-5}$, a weight decay of 0.03, and gradient clipping at 0.05. We sample with its guidance scale of 6.

\paragraph{Endora$^\dagger$.}
Endora~\cite{li2024endora} is an unconditional generator of $128^2$, 16-frame clips.
We extend it to a $288{\times}208$ canvas and add the first-frame latent and a mask as input channels. The first-frame latent is kept clean in the noisy input during training and at every denoising step.
We also add KB4 Pl\"ucker embeddings per patch through a linear layer.
Since no weights were released, we train it from scratch without image co-training, using the DINO prior weight of 0.1 from the code defaults instead of the 0.5 of the released training script.

\paragraph{FEAT$^\dagger$ and MedSora$^\dagger$.}
FEAT-L~\cite{wang2025feat} and MedSora-B~\cite{wang2024optical_medsora} are initialized from their released Colonoscopic weights.
FEAT$^\dagger$ receives the first frame in the same way as Endora$^\dagger$ and adds KB4 Pl\"ucker embeddings per patch.
MedSora$^\dagger$ takes the latent of the first frame and a mask as conditioning channels, together with the Pl\"ucker embeddings of the four frames in each latent. Its square architecture requires a $256{\times}256$ letterboxed canvas.
Neither model uses image co-training. MedSora$^\dagger$ omits the gradient clipping of the official code, which would not have been active since the gradient norm stayed below 0.4.

\paragraph{EndoGen$^\dagger$.}
EndoGen~\cite{liu2025endogen} is a class-conditional autoregressive model that tiles 16 frames of $128^2$ into a $4{\times}4$ grid and generates its $32{\times}32$ VQ tokens in raster order.
We initialize it from the released HyperKvasir weights, keep the VQ tokenizer frozen, and replace the class token with the null-class token.
We prepend the 64 tokens of the first frame as a prefix and also place them in the first tile, which is excluded from the loss.
The KB4 Pl\"ucker embedding of each $16{\times}16$\,px cell passes through a zero-initialized linear layer and is added to the input token that predicts that cell.

\paragraph{SurgSora$^\dagger$.}
SurgSora~\cite{chen2025surgsora} animates a frame with a frozen Stable Video Diffusion (SVD) model and a ControlNet.
The ControlNet takes RGB, depth, and semantic features of the first frame and warps them with a dense optical flow, which at test time is densified from trajectories sparsely drawn by the user.
We use SVD-XT 1.0, since the 1.1 weights used by the authors are gated. No trained ControlNet has been released, so we initialize ours from the SVD UNet.
We replace the trajectory with a dense camera flow. We obtain the flow by lifting the first frame of each chunk to 3D with monocular depth~\cite{yang2024depth}, moving it by the commanded poses, and reprojecting it with the KB4 model.
Following the original two-stage recipe, the ControlNet is first trained on optical flow from real videos for 2{,}000 updates. It is then trained on the camera flow for 14{,}000 updates, in place of the flow densified from sparse trajectories.
We rebuild the temporal fusion block for 25 frames and train in bf16 with standard AdamW, whereas the official second stage uses 8-bit Adam in fp16.
The official arguments specify a 500-update warm-up that the official constant schedule ignores, and we apply it.
We feed the VAE and ControlNet inputs in $[-1,1]$, as the official inference pipeline does.

\paragraph{EMA weights.}
Endora$^\dagger$, FEAT$^\dagger$, and MedSora$^\dagger$ are the three diffusion baselines trained with an exponential moving average (EMA).
For all three, the raw weights reach an equal or lower validation loss than the EMA weights (0.0882 vs.\ 0.1490, 0.1109 vs.\ 0.1110, and 0.1105 vs.\ 0.1113), so we generate with the raw weights instead of the EMA weights loaded by the official samplers.
For Endora$^\dagger$, trained from random initialization, the EMA still retains about 20\% of the initial weights after 16{,}000 updates.

\begin{table}[t]
\centering\footnotesize\setlength{\tabcolsep}{2pt}
\caption{Retrained baselines. Params counts the trained parameters, and Chunk is the number of frames in one training sample and one generation pass. UCPE$^\dagger$ trains on whole windows of 65 or 189 frames and generates the 189 frames in one pass.}
\label{tab:suppl_retrain}
\begin{tabular}{@{}l l r l@{}}
\toprule
Model & Initialization & Params & Chunk \\
\midrule
UCPE$^\dagger$     & backbone frozen, UCPE from zero         & 142M & window \\
Endora$^\dagger$   & from scratch                            & 678M & 16f \\
FEAT$^\dagger$     & Colonoscopic weights                    & 629M & 16f \\
MedSora$^\dagger$  & Colonoscopic weights                    & 87M  & 16f \\
EndoGen$^\dagger$  & HyperKvasir weights, VQ frozen          & 110M & 16f \\
SurgSora$^\dagger$ & SVD-XT 1.0 frozen, ControlNet from UNet & 710M & 25f \\
\bottomrule
\end{tabular}
\end{table}

\paragraph{Sanity checks.}
Before training, the denoisers of FEAT$^\dagger$ and MedSora$^\dagger$ reproduce the released models at their original $128^2$ resolution. The maximum difference is 0 for FEAT$^\dagger$ and $8.8{\times}10^{-5}$ for MedSora$^\dagger$, which comes from the recomputed position embedding.
The zero-initialized camera projection of EndoGen$^\dagger$ leaves the logits of the released model unchanged when the first-frame prefix is omitted.
The untrained UCPE branch outputs zero, and the zero-initialized output convolutions of the SurgSora$^\dagger$ ControlNet add nothing to the frozen SVD.
For SurgSora$^\dagger$, we warp the first frame of three validation clips with our camera flow, which moves the image in the commanded direction. The flow points outward from the focus of expansion for forward motion, and its mean horizontal component is $+49.5$\,px for a left turn and $-49.5$\,px for a right turn.

\section{Lumen Distance}
\label{sec:suppl_lumen}
The lumen distance in \cref{tab:ablation_cond} measures how closely the generated video follows the tube rendering.
We evaluate the 22 test clips at $256\times193$ within the image region of the first frame, using 64 uniformly spaced frames from frame 24 onward.
In each generated frame and in the tube rendering of the same frame, we take the darkest 10\% of pixels as the lumen and compute the distance between the two centroids, normalized by the image diagonal.
We average the distance over frames and the three seeds of each clip, and then over clips.
Without the tube, the distance increases in 19 of the 22 clips and in all four test patients.
Using the darkest 5\% or 20\% of pixels gives the same ordering of the test-time settings (b), (c), and Ours.

\paragraph{Lumen follows its own tube.}
A small lumen distance could also arise if the generated videos and the tube renderings simply share a typical lumen position.
We therefore compare each generated video with its own tube rendering and with the renderings of the other 21 test clips at the same frames.
We compute the IoU of the dark regions and define the gain as the IoU with the own rendering minus the mean IoU with the other renderings.
We also correlate the mean-centered lumen tracks of the generated video and its rendering, for the 21 clips whose rendered lumen moves by more than 0.02 of the diagonal.
The gain is 0.158 with the tube and 0.051 without it, and the track correlation is 0.52 and 0.05.
Both differences are significant under a two-sided sign-flip test over clips with Holm correction ($p<10^{-4}$).
Without the camera condition, the gain and the correlation stay close to ours, at 0.144 and 0.51.

\section{Evaluation Details}
\label{sec:suppl_eval}

\paragraph{Camera trajectory.}
We recover the camera trajectory of each generated video with MegaSaM~\cite{li2025megasam}, as for the training data.
All 189 frames are rectified from the KB4 model to $640{\times}480$ pinhole images with the fixed intrinsics of the clip, and MegaSaM runs at $512{\times}384$ with full bundle adjustment and without optimizing the intrinsics.
The poses are expressed relative to the first frame, with translations in units of its median depth.
The reference is the commanded trajectory, which is the MegaSaM trajectory of the real test clip.

\paragraph{Camera metrics.}
We follow the code of CamI2V~\cite{zheng2024cami2v}.
Each trajectory is cut into 9 windows of 16 frames sampled every 8 frames, starting at frames $0, 8, \dots, 64$, so that together they cover frames 0 to 184.
Each window is expressed relative to its first camera.
Within a window, the translations of each trajectory are divided by their largest distance from the first camera, and no further scale or similarity alignment is applied.
RotErr sums the geodesic angles between the generated and reference rotations, TransErr sums the distances between the camera positions, and CamMC sums the Frobenius norms of the differences between the $3{\times}4$ camera matrices.
We average each metric over the 9 windows and then over the 22 clips.
If MegaSaM fails on a generated video, the clip is scored as a static camera with identity poses.

\paragraph{Video quality.}
FVD uses the I3D network of StyleGAN-V on all 16-frame windows at a stride of 1, which gives 174 windows per clip, at $224{\times}224$ with the area outside the fisheye image masked.
The reference is the set of 22 real test clips.
FID uses the Inception network of StyleGAN2-ADA on all 189 frames of each clip, and IS uses the same network with 10 splits over all generated frames.

\paragraph{Full-procedure benchmark.}
The full-procedure tubes are synthetic, so no real video follows the same path.
The reference for FVD and FID is therefore real footage pooled from 12 held-out procedures.
For each generated path, we draw 1{,}914 I3D windows and 2{,}079 frames of clean reference footage with the same mix of colon segments as the path.
We compute the scores on 10 procedure-disjoint splits of the reference and average them over the splits and then over the six shapes.
On the generated side, we exclude the real start frame and every window that contains it.

\section{Camera Control without Pose Estimation on Generated Videos}
\label{sec:suppl_trackcheck}
The camera metrics in Tab.~\ref{tab:main} compare the commanded trajectory with the trajectory that MegaSaM recovers from each generated video.
To rule out a bias of the estimator on generated content, we also score camera control without estimating poses on the generated videos.
We track points on each generated test video with CoTracker3~\cite{karaev2025cotracker3} and compute the Sampson distance (Eq.~\ref{eq:sampson}) of the tracked correspondences at a frame gap of 16 under the commanded relative poses.
The frames are rectified from the KB4 model, and we keep the pairs whose median track residual after compensating the commanded rotation is at least 0.5\,px, as well as the pairs whose commanded motion is a pure rotation.
For each video, we take the median distance over the tracks of each pair and then over the pairs, and Tab.~\ref{tab:suppl_trackcheck} reports the median over videos.
Pass@2 is aggregated in the same way.
The real test clips, scored against their own commanded poses, give a floor of 1.62\,px.

\begin{table}[t]
\centering\footnotesize
\caption{Camera control without pose estimation on the generated videos. Median Sampson distance of tracked points under the commanded poses at frame gap 16, and Pass@2. MedSora$^\dagger$ is evaluated on 21 of the 22 clips. Best in \textbf{bold}.}
\label{tab:suppl_trackcheck}
\begin{tabular}{@{}l cc@{}}
\toprule
Method & Median (px)$\downarrow$ & Pass@2$\uparrow$ \\
\midrule
Cosmos3-Nano 9D & 15.15 & 0.05 \\
Wan2.2-Fun-5B-Control-Camera & 13.54 & 0.05 \\
CamI2V & \phantom{0}7.41 & 0.12 \\
LingBot-World Base-Cam & 19.54 & 0.04 \\
UCPE$^\dagger$ & 14.30 & 0.06 \\
\midrule
Endora$^\dagger$ & 27.98 & 0.00 \\
FEAT$^\dagger$ & 12.26 & 0.03 \\
MedSora$^\dagger$ & 23.72 & 0.02 \\
EndoGen$^\dagger$ & 15.99 & 0.04 \\
SurgSora$^\dagger$ & 22.83 & 0.03 \\
\midrule
\methodname{} & \phantom{0}\textbf{3.39} & \textbf{0.30} \\
\midrule
Real test clips & \phantom{0}1.62 & 0.58 \\
\bottomrule
\end{tabular}
\end{table}

\methodname{} has the lowest distance and the highest Pass@2 among all generators.
With the same seed, it has a lower distance than UCPE$^\dagger$ on all 22 clips (Wilcoxon signed-rank test, $p=4.8\times10^{-7}$).
Scoring the videos of \methodname{} against the commanded poses of a different clip raises its distance to 24.7\,px, so the score reflects whether the generated motion follows the command.
Each of the three seeds of a clip counts as one video, except for the $^\dagger$ baselines, which use one seed as in Tab.~\ref{tab:main}.

\end{document}